\documentclass{article}

\usepackage[utf8]{inputenc}
\usepackage[T1]{fontenc}
\usepackage{CJKutf8}
\usepackage{newunicodechar}
\newunicodechar{，}{,}

\usepackage{longcat_style}

\usepackage{amsmath}
\usepackage{amsfonts}
\usepackage{amssymb}
\usepackage{bbm}
\usepackage{nicefrac}

\usepackage{xcolor}

\usepackage{array}
\usepackage{booktabs}
\usepackage{tabularx}
\usepackage{multirow}
\usepackage{makecell}
\usepackage{colortbl}
\usepackage{arydshln}

\usepackage{graphicx}
\usepackage{svg}
\usepackage{wrapfig}
\usepackage{subcaption}

\usepackage{enumitem}
\usepackage{microtype}
\usepackage{lipsum}
\usepackage{siunitx}
\usepackage{xspace}

\usepackage{url}
\usepackage{xurl}
\usepackage{natbib}
\usepackage{doi}

\usepackage{pifont}
\usepackage[most]{tcolorbox}

\usepackage{hyperref}
\usepackage{cleveref}

\hypersetup{
    colorlinks=true,
    linkcolor=blue,
    urlcolor=blue,
    citecolor=blue,
    linkbordercolor=blue,
    urlbordercolor=blue,
    citebordercolor=blue,
    pdfborderstyle={/S/U/W 1},
}

\setlist[itemize]{leftmargin=*}
\setlist[enumerate]{leftmargin=*}
\setlist[description]{leftmargin=*}

\definecolor{mygray}{gray}{.88}
\definecolor{mycyan}{cmyk}{.15,0,0,0}
\definecolor{mycyan2}{cmyk}{.85,0,0,0}
\definecolor{mygreen}{rgb}{0.19, 0.79, 0.02}
\definecolor{midnightgreen}{rgb}{0.0, 0.29, 0.33}
\definecolor{darkgreen}{RGB}{0,160,0}

\newcommand{\notcheckmark}{\textcolor{black}{\bcmark\kern-1.1ex\raisebox{.7ex}{\rotatebox[origin=c]{125}{--}}}\color{black}}
\newcommand{\bcmark}{\color{blue}{\ding{51}}}

\usepackage{tikz}
\usepackage{amsthm}
\usepackage{mdframed}

\usepackage[utf8]{inputenc} % allow utf-8 input
\usepackage{csquotes}

\usepackage{ulem}
\usepackage{caption}
\usepackage{multibib}
\usepackage{color}

\usepackage{float}
\usepackage{inconsolata}

\tcbuselibrary{listings,theorems}
\newtcolorbox{mybox}{colback=white!5!white,colframe=black!75!black, left=.05in, right=.05in}

\definecolor{bluex}{rgb}{0.27, 0.42, 0.81}
\definecolor{purplex}{HTML}{9564bf}
\definecolor{red3}{HTML}{C52A20}
\definecolor{red2}{HTML}{B36A6F}
\definecolor{red1}{HTML}{FFb5b5}
\definecolor{purple}{HTML}{B36A6F}
\definecolor{darkyellow}{HTML}{D5BA82}
\definecolor{blue1}{HTML}{508AB2}
\definecolor{blue2}{HTML}{C4E4E3}
\definecolor{green1}{HTML}{A1D0C7}
\definecolor{green2}{HTML}{BFF6BA}
\definecolor{green3}{HTML}{028100}
\definecolor{teal}{HTML}{508AB2}
\definecolor{purple1}{HTML}{8d3a94}

\newtcbtheorem[]{exmp}{Example}%
{colback=purple1!5,colframe=purple1!80,fonttitle=\bfseries, left=.08in, right=.08in,bottom=.05in, top=.05in}{exmp}

\title{\textsc{AMO}-Bench: Large Language Models Still\\Struggle in High School Math Competitions}

\author{
Shengnan An\thanks{\, Correspondence to: \texttt{\{anshengnan, caoxuezhi\}@meituan.com}.}\,\,$^{\diamondsuit}$, Xunliang Cai$^{\diamondsuit}$, Xuezhi Cao$^{* \diamondsuit}$, Xiaoyu Li$^{\diamondsuit}$, Yehao Lin$^{\diamondsuit}$, Junlin Liu\thanks{\, Work done during the internship at Meituan.}\,\,$^{\clubsuit}$,\\ \,\textbf{Xinxuan Lv$^{\diamondsuit}$, Dan Ma$^{\diamondsuit}$, Xuanlin Wang$^{\dagger\heartsuit}$, Ziwen Wang$^{\diamondsuit}$, Shuang Zhou$^{\diamondsuit}$}\\(Alphabetical order by last name)\\
\\
$^{\diamondsuit}$Meituan\quad
$^{\clubsuit}$University of Chinese Academy of Sciences\,\,\,
$^{\heartsuit}$Harbin Institute of Technology
}

\renewcommand{\headeright}{\raisebox{-0.2\height}{\includegraphics[height=1.8em]{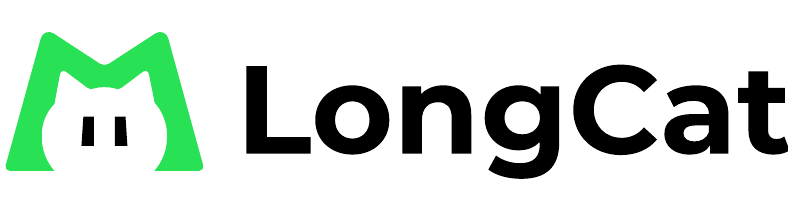}}}
\renewcommand{\shorttitle}{\small \textsc{AMO}-Bench: Large Language Models Still Struggle in High School Math Competitions}

\begin{document}
\maketitle

\begin{abstract}
We present \textsc{AMO}-Bench, an \textbf{A}dvanced \textbf{M}athematical reasoning benchmark with \textbf{O}lympiad level or even higher difficulty, comprising 50 human-crafted problems.
Existing benchmarks have widely leveraged high school math competitions for evaluating mathematical reasoning capabilities of large language models (LLMs).
However, many existing math competitions are becoming less effective for assessing top-tier LLMs due to performance saturation (e.g., AIME24/25).
To address this, \textsc{AMO}-Bench introduces more rigorous challenges by ensuring all 50 problems are (1) cross-validated by experts to meet at least the International Mathematical Olympiad (IMO) difficulty standards, and (2) entirely original problems to prevent potential performance leakages from data memorization.
Moreover, each problem in \textsc{AMO}-Bench requires only a final answer rather than a proof, enabling automatic and robust grading for evaluation.
Experimental results across 26 LLMs on \textsc{AMO}-Bench show that even the best-performing model achieves only 52.4\% accuracy on \textsc{AMO}-Bench, with most LLMs scoring below 40\%.
Beyond these poor performances, our further analysis reveals a promising scaling trend with increasing test-time compute on \textsc{AMO}-Bench.
These results highlight the significant room for improving the mathematical reasoning in current LLMs.
We release \textsc{AMO}-Bench to facilitate further research into advancing the reasoning abilities of language models.
\end{abstract}

\begin{center}
\textbf{Code, Dataset, and Leaderboard:} { } \href{https://amo-bench.github.io}{\texttt{amo-bench.github.io}}
\end{center}

\begin{figure}[ht]
    \centering
    \includegraphics[width=0.89\textwidth]{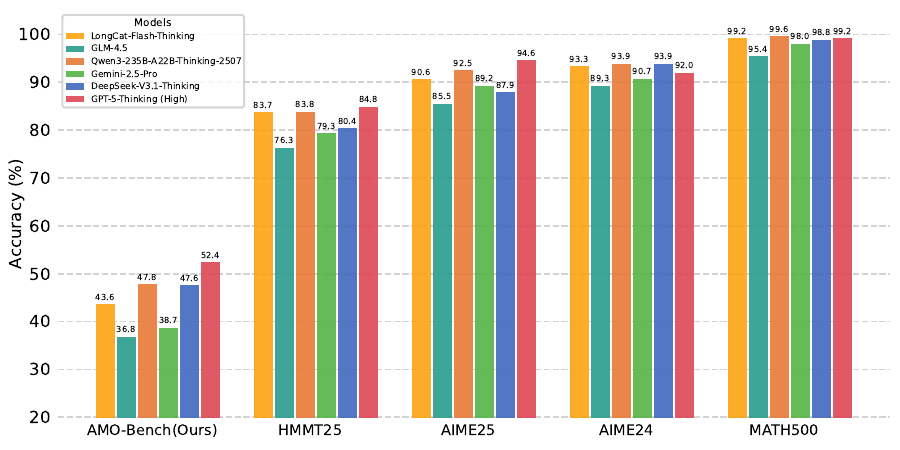} % 替换为你的文件名
    \caption{Performance of top-tier reasoning models on \textsc{AMO}-Bench as well as existing competition-level math benchmarks.
    Except for the results on \textsc{AMO}-Bench, all other results are sourced from \citet{longcatflashthinking}.}
    \label{fig:cover}
\end{figure}

\section{Introduction}

Recent advances in large language models (LLMs) have demonstrated significant improvements in reasoning capabilities~\citep{o1,gemini25,gpt5,claude4,grok4,qwen3,deepseekr1,deepseekv3,longcatflash,glm,seed15,hunyuan,kimik2,longcatflashthinking}.
To track this rapid progress, mathematical problem solving has become a critical metric for evaluation, as it inherently demands complex and multi-step reasoning processes to arrive at correct answers.
As a result, many current benchmarks utilize problems from high school mathematics competitions (e.g., HMMT and AIME) to assess the reasoning abilities of LLMs~\citep{balunovic2025matharena,he2024olympiadbench,gaoomni,fang2025mathodyssey}.
Recent results indicate that state-of-the-art models are achieving remarkable performances on these benchmarks, with some even surpassing 90\% accuracy on competitions like AIME24/25.

However, these impressive results also expose an emerging challenge: many existing mathematics benchmarks are approaching performance saturation and are becoming less effective for assessing further advancements in reasoning capabilities.
On the one hand, as LLMs gradually approach or even surpass human-level capabilities in mathematics, some math competitions are becoming less challenging for top-tier models~\citep{gpt5,deepseekv3,qwen3,longcatflashthinking}.
On the other hand, most current benchmarks are derived from previous competitions, raising concerns about potential data memorization and performance leakage~\citep{sun2025challenging,balunovic2025matharena}.
While recent efforts have incorporated problems from more difficult and newly held contests such as the International Mathematical Olympiad (IMO), these questions tend to be proof-based and require manual verification by experts~\citep{balunovic2025matharena,usamo}.
This reliance on expert review hinders the implementation of automated scoring processes, leading to inefficiency and inconsistency in large-scale evaluations and result reproductions.

To address these limitations, we present \textsc{AMO}-Bench, an advanced mathematical reasoning benchmark consisting of 50 novel and extremely challenging problems.
The core features of \textsc{AMO}-Bench are as follows:
\begin{itemize}
    \item \textbf{Original problems.} To prevent performance leaks from existing resources as much as possible, all problems in \textsc{AMO}-Bench are newly crafted by human experts.
    Moreover, we conduct a secondary verification to ensure that there are no highly similar problems in existing competitions or online resources.
    \item \textbf{Guaranteed difficulty.} Each problem has undergone rigorous cross-validation by multiple experts to ensure it meets at least the difficulty standards of IMO.
    We also incorporate an LLM-based difficulty filtering stage to exclude questions that do not present sufficient challenge to current reasoning models.
    \item \textbf{Final-answer based grading.} Each problem in \textsc{AMO}-Bench requires a final answer rather than a full proof, enabling efficient automatic grading.
    For each problem, we employ a parser-based or LLM-based grading method according to its answer type, balancing the grading cost and generalizability. 
    \item \textbf{Human-annotated reasoning paths.} In addition to the final answer, each problem also includes a detailed reasoning path written by human experts.
    These additional annotations enhance solution transparency and could support further explorations on \textsc{AMO}-Bench, such as prompt engineering and error analysis.
\end{itemize}

% results and analysis
Experimental results across various LLMs demonstrate that contemporary LLMs still struggle with the significant challenges presented by \textsc{AMO}-Bench.
Among 26 evaluated models, the state-of-the-art accuracy on \textsc{AMO}-Bench is only 52.4\%, achieved by GPT-5-Thinking (High), with most models scoring below 40\%.
Figure~\ref{fig:cover} illustrates the performance of several leading models on \textsc{AMO}-Bench as well as the comparison with other mathematical benchmarks.
Beyond their limited final performances on \textsc{AMO}-Bench, LLMs consume substantially more output tokens in \textsc{AMO}-Bench compared to existing evaluation datasets.
For example, GPT-5-Thinking (High) generates an average of approximately 37K output tokens for \textsc{AMO}-Bench, whereas it produces only about 7K and 6K tokens for AIME25 and AIME24, respectively.
This exceptionally high token consumption further underscores the difficulty of \textsc{AMO}-Bench for current LLMs.
Despite the poor performances of current LLMs, our analysis also reveals considerable potential for further improvements.
Notably, top-tier models achieve pass@32 rates exceeding 70\%, suggesting they possess the initial capability to solve these challenging problems even if they do not consistently identify the correct reasoning path at present.
Furthermore, we show that the model performances exhibit a near-linear growth trend relative to the logarithm of output length, indicating continued benefits from test-time scaling.
These analyses suggest substantial opportunities remain to enhance reasoning capabilities in future generations of language models.

The data and evaluation code of \textsc{AMO}-Bench are publicly available at \href{https://amo-bench.github.io}{\texttt{amo-bench.github.io}}.
We hope this novel and challenging benchmark will facilitate further research into advancing the reasoning abilities of language models.

\section{\textsc{AMO}-Bench}\label{sec:amo_bench}

\begin{figure}[t]
    \centering
    \includegraphics[width=0.99\textwidth]{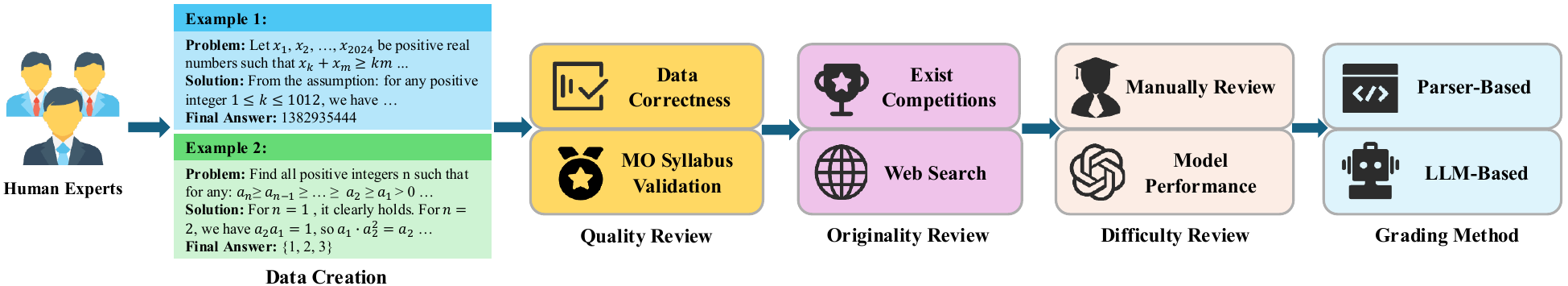} 
    \caption{The construction and grading pipeline of \textsc{AMO}-Bench.}
    \label{fig:pipeline}
\end{figure}

In this section, we first introduce the construction process of \textsc{AMO}-Bench (Section~\ref{sec:data_construct}) and present the basic statistics of this dataset (Section~\ref{sec:data_stat}).
Then, we elaborate on the grading methodology designed for \textsc{AMO}-Bench (Section~\ref{sec:grading}).
Figure~\ref{fig:pipeline} briefly illustrate the construction and grading pipeline of \textsc{AMO}-Bench.

\subsection{Construction Pipeline}\label{sec:data_construct}

To ensure the high standards of quality, originality, and difficulty level in our dataset, we have built up a comprehensive multi-stage construction pipeline that covers the entire process from question creation to final inclusion.
This pipeline comprises four major stages: data creation, quality review, originality review, and difficulty review.

\paragraph{Data creation.}
All problems are independently designed by mathematics experts from top universities and educational institutions.
These experts have extensive backgrounds in high school mathematics competitions, either having won MO-level mathematics competition awards or possessing experience in competition problem design.
Beyond the final answer, each problem author must provide a detailed step-by-step solution.
These annotated solutions will be utilized in the subsequent quality review stage and will also aid in assessing the overall difficulty of \textsc{AMO}-Bench (see Section~\ref{sec:data_stat} for details).

\paragraph{Quality review.}
Each candidate problem undergoes blind review by at least three experts to assess its quality.
This quality review stage focuses primarily on two aspects:
\begin{itemize}
    \item Whether the problem statement and solution are semantically unambiguous and logically correct.
    \item Whether the mathematical knowledge required for the problem is within the scope typically covered in MO-level competitions such as IMO.
\end{itemize}

\paragraph{Originality review.}
The originality review stage aims to ensure that these newly created problems are not mere rewrites of publicly available materials, but demonstrate genuine originality.
To this end, we assess the originality of each problem through the following methods:
\begin{itemize}
    \item Compare it against problems in existing datasets (e.g., AIME24/25) with 10-gram matching.
    \item Conduct web searches to identify any similar online content.
\end{itemize}
Additionally, during the quality review stage, experts are also required to indicate whether they have encountered highly similar questions in past competitions.

\paragraph{Difficulty review.}
To ensure that \textsc{AMO}-Bench presents a sufficient challenge to state-of-the-art LLMs, we implement a difficulty review stage to filter out problems lacking adequate complexity (even if they may be suitable for some MO-level competitions, e.g., the first 10 questions in AIME).
Specifically, each selected problem must satisfy the following two criteria:
\begin{itemize}
    \item The problem must meet or exceed the IMO difficulty standards, as verified by the human expert.
    \item We employed multiple advanced reasoning models (such as GPT, DeepSeek, and Gemini series models) for preliminary evaluation, requiring that at least two such models fail to correctly and consistently solve the problem\footnote{For each model, our preliminary evaluation involves three samples. If all three samples are correct, the model is deemed capable of consistently solving the problem.}.
\end{itemize}

\subsection{Dataset Statistics}\label{sec:data_stat}

\begin{figure}[t]
  \centering
  \begin{subfigure}[b]{0.39\textwidth}
    \centering
    \includegraphics[width=0.8\textwidth]{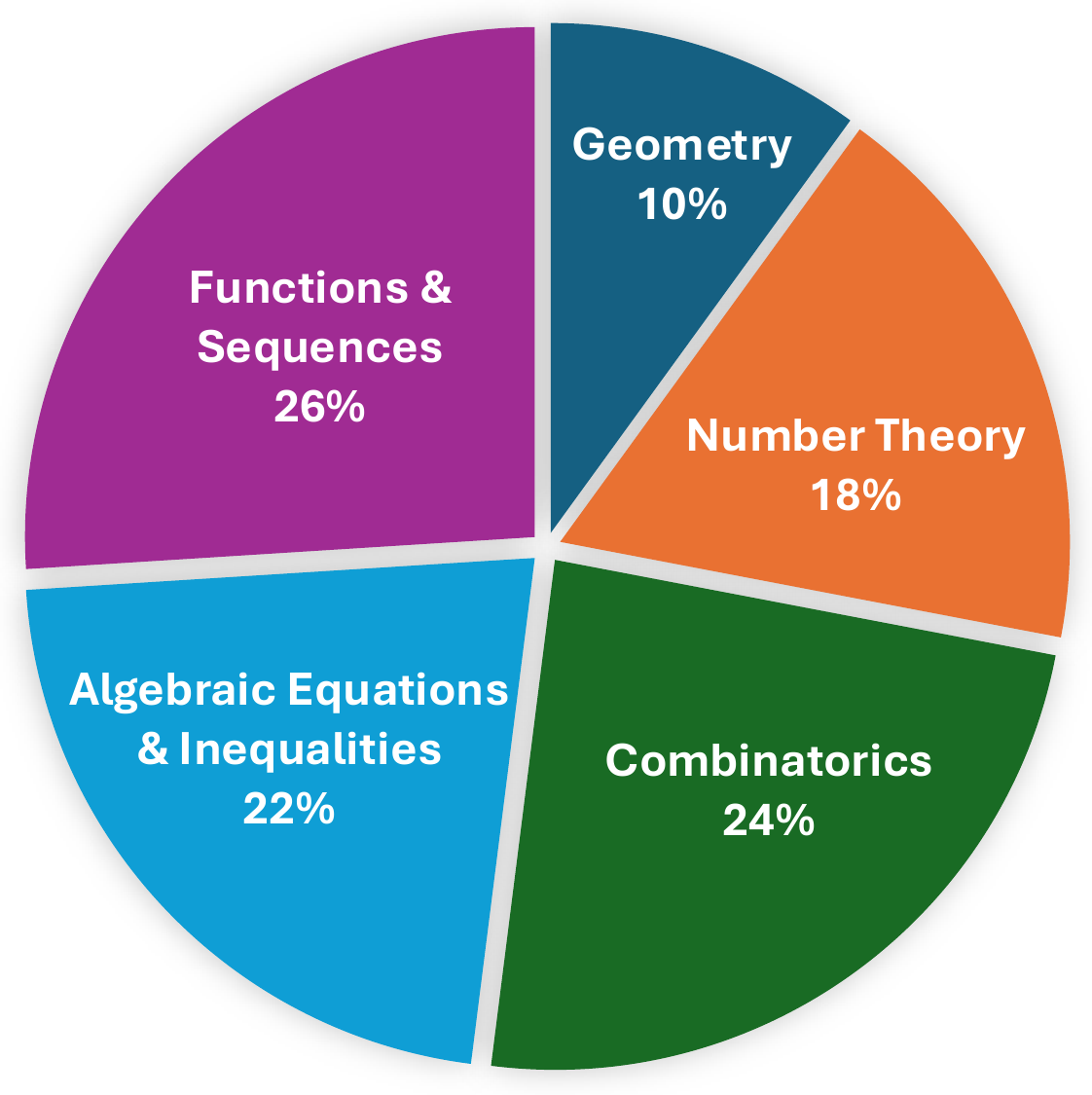}
    \caption{Distribution of problem categories.}
    \label{fig:category}
  \end{subfigure}
  \hfill
  \begin{subfigure}[b]{0.59\textwidth}
    \centering
    \includegraphics[width=0.99\textwidth]{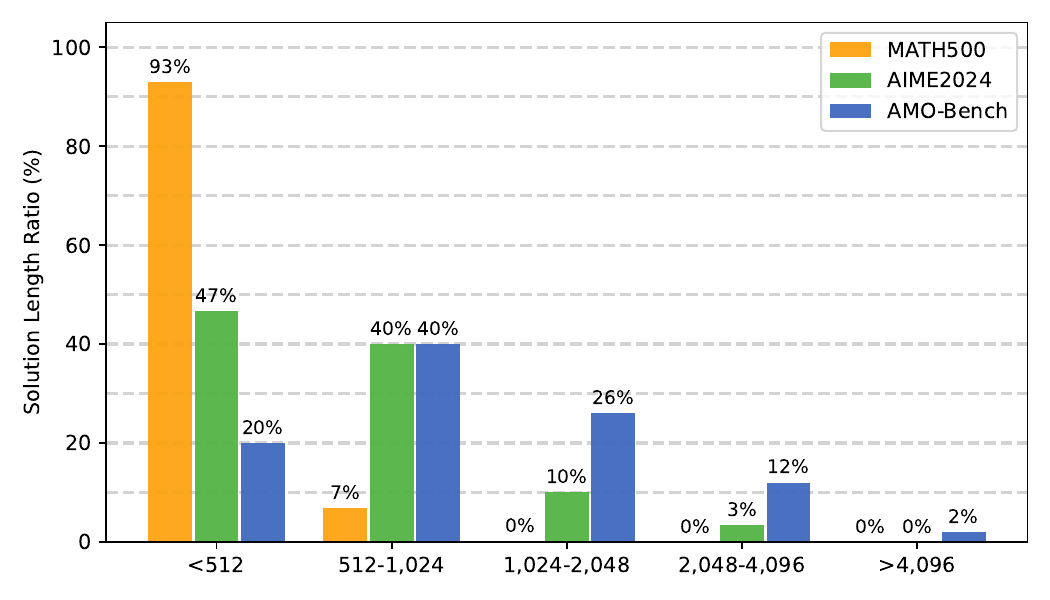}
    \caption{Comparison of solution lengths.}
    \label{fig:length}
  \end{subfigure}
  \caption{Basic statistics of \textsc{AMO}-Bench. (a) The distribution of problem categories in \textsc{AMO}-Bench. (b) The distribution of human-annotated solutions in \textsc{AMO}-Bench as well as the comparison with MATH500 and AIME24.}
  \label{fig:two_images_sub}
\end{figure}

\paragraph{Problem categories.} 
Referring several official competition syllabus, we categorize the 50 problems of \textsc{AMO}-Bench into the following five primary categories:
Algebraic Equations \& Inequalities (11/50), Functions \& Sequences (13/50), Geometry (5/50), Number Theory (9/50), and Combinatorics (12/50).
Figure~\ref{fig:category} show the overall distribution of problem categories in \textsc{AMO}-Bench.

\paragraph{Length distribution of human-annotated solutions.}
Since the problems in our \textsc{AMO}-Bench are equipped with manually annotated solutions, we can preliminarily analyze the reasoning complexity of these problems from the view of solution length.
We measure solution length in terms of token count\footnote{We use the tokenizer of DeepSeek-V3.1 model to count tokens in solutions.}.
Additionally, we compare the distribution of solution lengths with those from AIME24\footnote{\href{https://huggingface.co/datasets/HuggingFaceH4/aime_2024}{https://huggingface.co/datasets/HuggingFaceH4/aime\_2024}.} and MATH500\footnote{\href{https://huggingface.co/datasets/HuggingFaceH4/MATH-500}{https://huggingface.co/datasets/HuggingFaceH4/MATH-500}.}.
Figure~\ref{fig:length} illustrates the solution length distributions across these benchmarks.
It reveals that solutions in \textsc{AMO}-Bench exhibit significantly higher lengths, indicating that problems in this benchmark are inherently more challenging and require more complex reasoning to arrive at the final answer.
We conduct a further analysis of the model solution lengths in Section~\ref{sec:exp_results}.

\subsection{Grading Method}\label{sec:grading}
For evaluating answers generated by LLMs, prior work has primarily utilized two approaches: parser-based grading and LLM-based grading.
Parser-based grading offers high efficiency and accuracy when the model's response can be successfully parsed; however, its applicability is limited to simple answer formats such as numerical values or sets, making it challenging to assess more complex answers.
In contrast, LLM-based grading provides greater flexibility across diverse answer types but may be less efficient and does not consistently guarantee accuracy.

To fully leverage the strengths of both grading methods, \textsc{AMO}-Bench employs different grading approaches based on the specific answer type for each problem.
Specifically, problems in \textsc{AMO}-Bench are divided into four main answer types: numerical answers (e.g., Example~\ref{exmp:numerical_answer}), set answers (e.g., Example~\ref{exmp:set_answer}), variable-expression answers (e.g., Example~\ref{exmp:variable_answer} which requires providing the general formula for an arithmetic sequence), and descriptive answers (e.g., Example~\ref{exmp:descriptive_answer} which involves comprehensively considering multiple scenarios).
The prompt templates for used for grading are contained in Appendix~\ref{sec:prompts}.

\begin{exmp}{Problem with Numerical Answer}{numerical_answer}
{}\textbf{Question:} Let \({x}_{1},{x}_{2},\cdots ,{x}_{2024}\) be positive real numbers such that \({x}_{k} + {x}_{m} \geq {km}\) for any \(1 \leq k < m \leq {2024}\). Find the minimum value of \({x}_{1} + {x}_{2} + \cdots + {x}_{2024}\). \par
\textbf{Answer:} \boxed{1382935444}
\end{exmp}

\begin{exmp}{Problem with Set Answer}{set_answer}
{}\textbf{Question:} Find all positive integers \(\mathbf{n}\) such that for any: \({a}_{n} \geq {a}_{n - 1} \geq {a}_{n - 2} \geq \cdots \cdots {a}_{2} \geq {a}_{1} > 0,\) satisfying \(\mathop{\sum }\limits_{{k = 1}}^{n}{a}_{k} = \mathop{\sum }\limits_{{k = 1}}^{n}\frac{1}{{a}_{k}}\), the inequality \(\mathop{\prod }\limits_{{k = 1}}^{n}{a}_{k}^{k} \geq 1\) holds. \par
\textbf{Answer:} \boxed{\{1, 2, 3\}}
\end{exmp}

\begin{exmp}{Problem with Variable-Expression Answer}{variable_answer}
{}\textbf{Question:} The sequence $\{a_n\}_{n=1}^{\infty}$ consists of positive terms, with $a_1=7$, $a_2=2$, and satisfies the recurrence relation
$$8a_{n+2}^4=3+4a_{n+1}+a_n\quad(n\in\mathbb{N}^*).$$
Find the general term formula for this sequence. \par
\textbf{Answer:} $$\boxed{\dfrac{(2+\sqrt{3})^{2^{2-n}}+(2-\sqrt{3})^{2^{2-n}}}{2}}$$
\end{exmp}

\begin{exmp}{Problem with Descriptive Answer}{descriptive_answer}
{}\textbf{Question:} Let \(n\) be an integer with \(n>2\).
Real numbers \(a_1, a_2, \ldots, a_n\) satisfy
\[
\sum_{k=1}^{n} a_k = 2n, \qquad \sum_{k=1}^{n} k\,|a_k| = 4n.
\]Find the minimum value of \(a_1^2 + a_2^2 + \cdots + a_n^2\). \par
\textbf{Answer:} For \(n=3\), the minimum of \(a_1^2 + a_2^2 + a_3^2\) is \(12\). \par
~~~~~~~~~~~~~~~~For \(n \ge 4\), the minimum of \(a_1^2 + a_2^2 + \cdots + a_n^2\) is \(\dfrac{6n^2}{5}\).
\end{exmp}

For problems requiring numerical, set, or variable-expression answers (39 out of 50), we employ the parser-based grading.
The evaluated LLMs are instructed to format their final responses as \texttt{\textbackslash boxed\{<answer>\}}.
We then utilize the tools provided by \texttt{math-verify}\footnote{\href{https://github.com/huggingface/Math-Verify}{https://github.com/huggingface/Math-Verify}.} to parse these answers and verify the equivalence with the ground truth.
Moreover, if the model answer containing decimal values, we require an accuracy of at least four decimal places.
For variable-expression answers, we assign multiple sets of values to the variables in the expression, then verify whether the values of the generated expression match that of the ground-truth expression.
We also manually review the parsing results during the preliminary evaluation and adjust the post-processing algorithms.
    
For problems requiring descriptive answers (11 out of 50), we use LLM-based grading with o4-mini (Low) serving as the grading model.
To ensure robust assessment, majority voting is performed across five independent grading samples for each response.
Additionally, during preliminary evaluation, we manually verify the correctness of LLM-based grades for all descriptive answers and revise answer descriptions where needed to enhance grading accuracy.

\paragraph{Grading accuracy.}
Prior to conducting the large-scale evaluation, we performed a manual quality check to ensure the reliability of the designed grading method.
This assessment included 1,000 responses generated by 10 different LLMs.
The results indicate that the grading accuracy reached 99.2\%, providing strong validation for the effectiveness of the grading method on \textsc{AMO}-Bench.

\section{Experiments}

In this section, we present the experimental results on \textsc{AMO}-Bench.
We first describe the experimental setup (Section~\ref{sec:setup}), followed by a discussion of the main results and analysis (Section~\ref{sec:exp_results}).

\begin{figure}[t]
    \centering
    \includegraphics[width=0.99\textwidth]{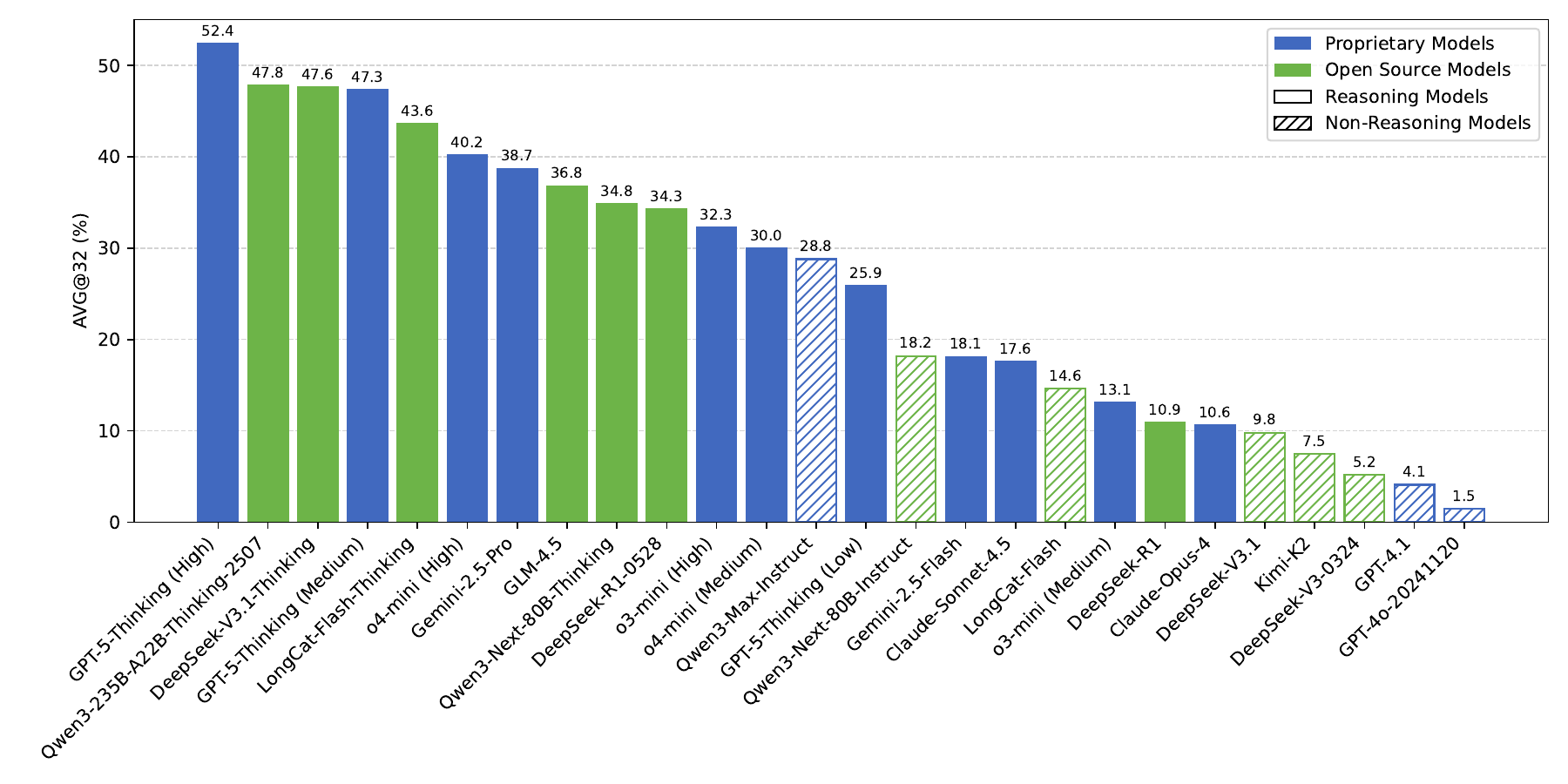} % 替换为你的文件名
    \caption{The AVG@32 performance of various LLMs on \textsc{AMO}-Bench.}
    \label{fig:main_result}
\end{figure}

\subsection{Experimental Setup}\label{sec:setup}

\paragraph{Models.}
To conduct a comprehensive and representative evaluation on \textsc{AMO}-Bench, we select a diverse set of leading LLMs, encompassing both open-source models and proprietary models.
Specifically, the evaluation includes top-tier models provided by OpenAI~\citep{gpt5}, Gemini~\citep{gemini25}, Anthropic~\citep{claude4}, DeepSeek~\citep{deepseekr1}, Qwen~\citep{qwen3}, GLM~\citep{glm}, Moonshot~\citep{kimik2}, and LongCat~\citep{longcatflashthinking}.
In addition to evaluating reasoning models that have been specifically enhanced for long-term thinking tasks, we also incorporated several powerful non-reasoning models to demonstrate their potential in tackling complex reasoning challenges.

\paragraph{Sampling settings.}
We set the \texttt{temperature} of sampling to 1.0 for reasoning models and 0.7 for non-reasoning models.
For all evaluated models, we use \texttt{top-k}=50 and \texttt{top-p}=0.95 during sampling.
We configure the maximum context/output length to the highest allowable limit for each model during inference.
This avoids underestimating the reasoning capabilities of the model due to restrictions on the token budget.
To ensure the stability of the final evaluation results, we sampled the results from each model 32 times and reported the average performance of these 32 results as the final metric (denoted as AVG@32).
Appendix~\ref{sec:sample_num} illustrates the fluctuation of the average result across different sampling times. It demonstrates that when sampling 32 times, the average model performance exhibits a relatively small fluctuation and rarely appears to reverse the model ranking order.

\subsection{Results and Analysis}~\label{sec:exp_results}
\begin{figure}[t]
    \centering
    \includegraphics[width=0.99\textwidth]{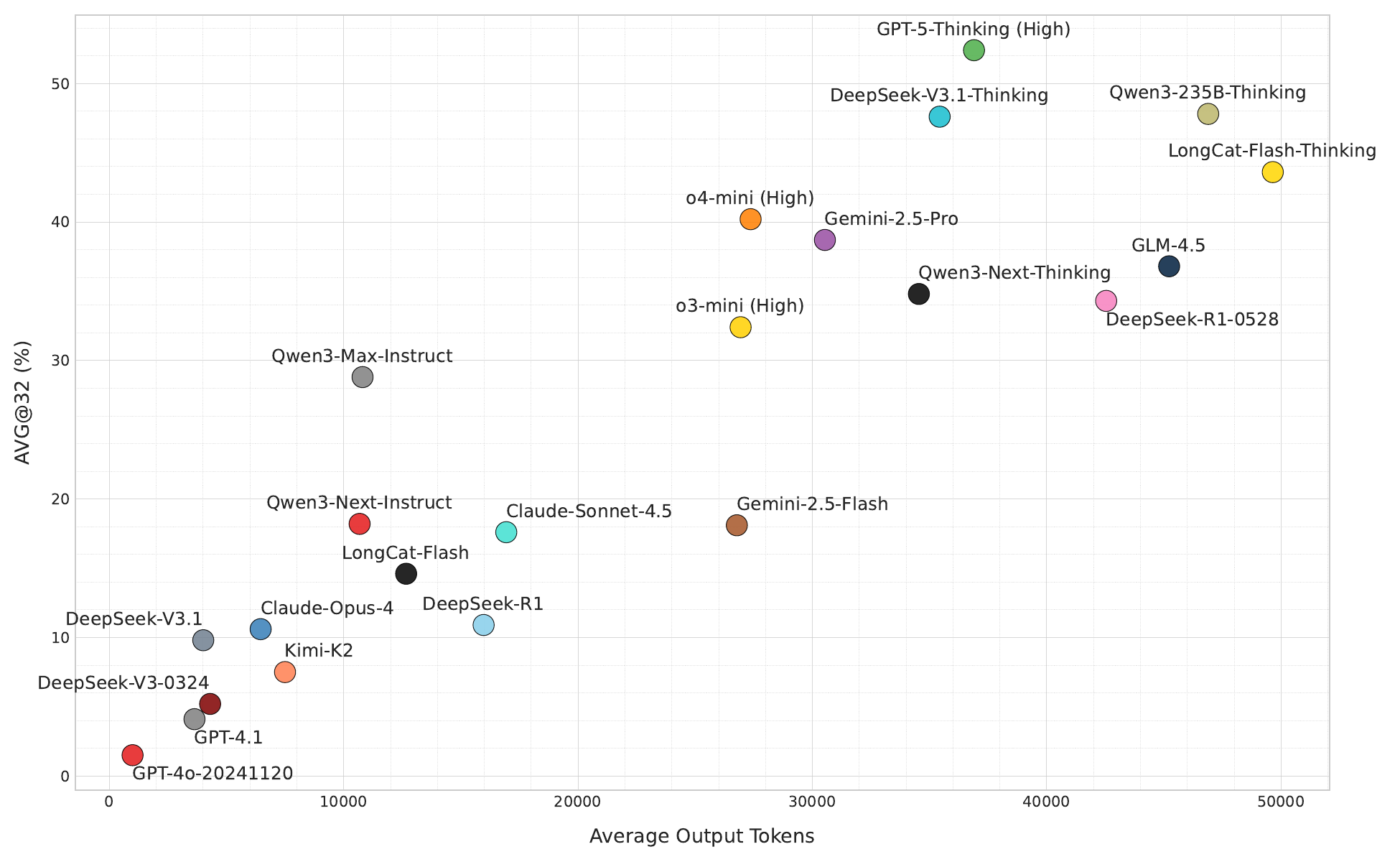} 
    \caption{The AVG@32 performance of LLMs vs. the average model output length.}
    \label{fig:acc_output_length}
\end{figure}

\paragraph{Main results.}
Figure~\ref{fig:main_result} presents the AVG@32 performance of various leading LLMs, categorized by proprietary/open-source status and reasoning/non-reasoning properties\footnote{To facilitate easier reproduction and utilization of \textsc{AMO}-Bench, you can take a fast try on the \textsc{AMO}-Bench-P subset, which includes only the 39 parser-based grading problems from \textsc{AMO}-Bench.
Appendix~\ref{sec:amo_bench_p} presents the AVG@32 performance of LLMs on \textsc{AMO}-Bench-P.}.
Overall, all these models still struggle with the significant challenges presented by \textsc{AMO}-Bench.
Even the highest performing model GPT-5-Thinking (High) reaches just 52.4\%, while most others score below 40\%.
This indicates substantial room for improvement in complex reasoning abilities across all current language models.
Moreover, both proprietary and open-source reasoning models occupy top ranks in the leaderboard, indicating that recent open-source advancements are closing the gap with leading commercial models.
The best-performing open-source model is only about 5\% lower than the top proprietary result.
Besides reasoning models, some non-reasoning models demonstrate a performance exceeding expectations, such as Qwen3-Max-Instruct and LongCat-Flash.
These non-reasoning models even outperforms several reasoning models such as o3-mini (Medium), indicating their significant potential in tackling complex reasoning tasks.

\paragraph{Comparison of reasoning efficiency.}
Figure~\ref{fig:acc_output_length} shows the average output length and the AVG@32 performance of each model.
Overall, it demonstrates a clear trend that higher-performing models tend to require more output tokens.
The first-tier models that reach higher than 40\% AVG@32 scores utilize more than 35K completion tokens.
Even among non-reasoning models, those with superior performance are distinguished by their ability to process more tokens, sometimes reaching levels comparable to reasoning models.
Additionally, when examining models within the same series, there are notable improvements in reasoning efficiency over time.
For example, o4-mini (High) outperforms o3-mini (High) at similar or slightly increased token counts.
Likewise, DeepSeek-V3.1-Thinking shows significant gains compared to DeepSeek-R1-0528 with even significantly less output tokens.

\begin{figure}[t]
    \centering
    \includegraphics[width=0.99\textwidth]{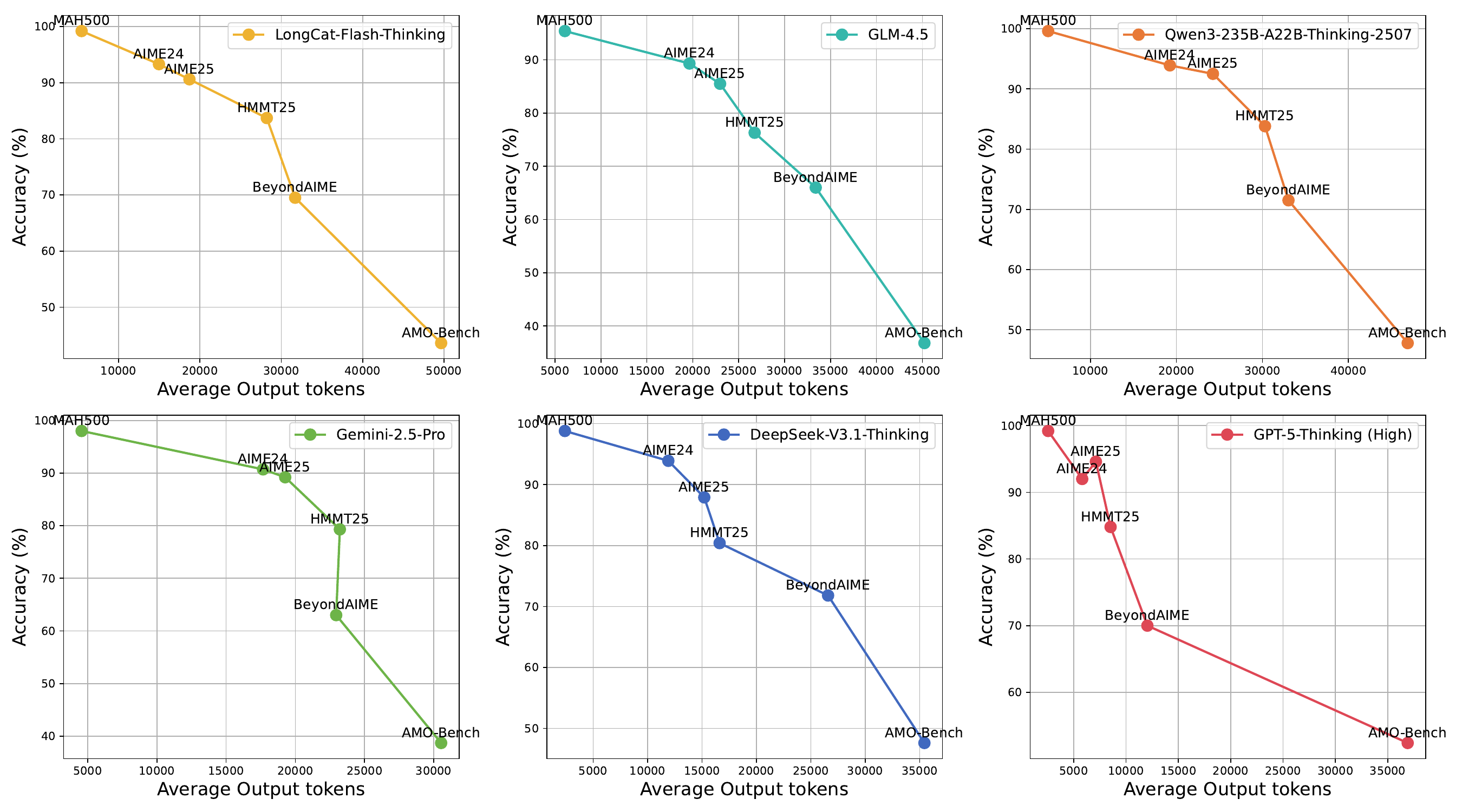}
    \caption{The relationship between accuracy and average output length on different math benchmarks.}
    \label{fig:bmk_tokens}
\end{figure}

\begin{figure}[t]
    \centering
    \includegraphics[width=0.95\textwidth]{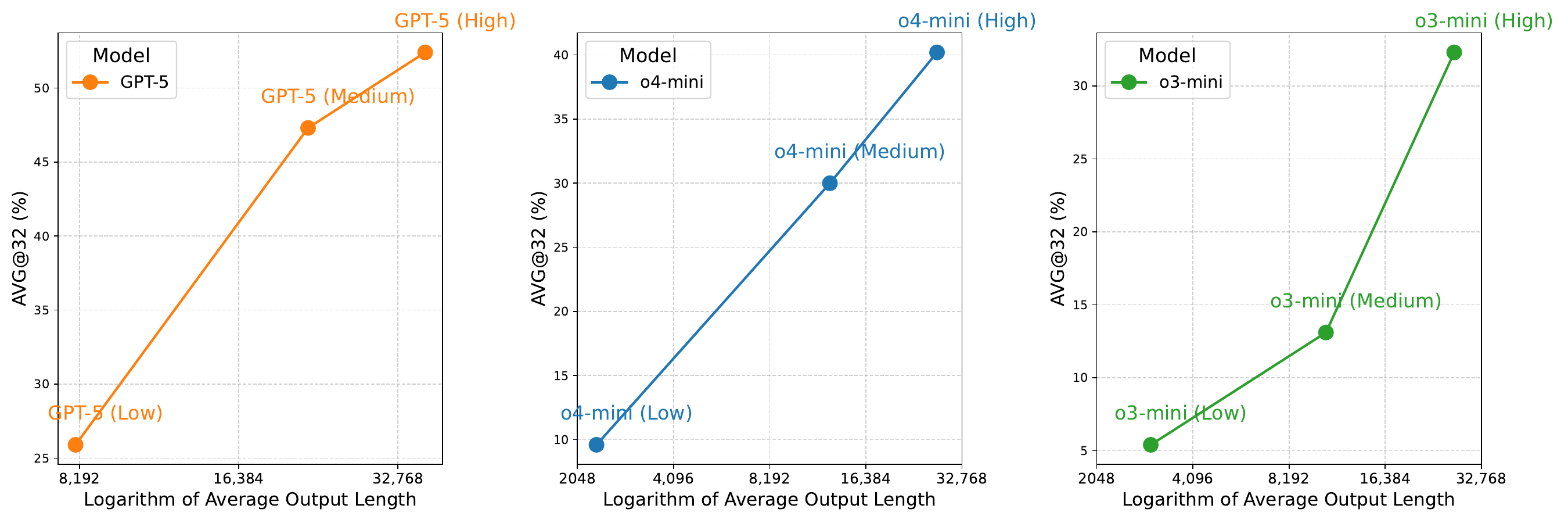}
    \caption{The model performance and output length under different reasoning effort settings.}
    \label{fig:reasoning_effort}
\end{figure}

Beyond the main results outlined above, we also provide further analysis and insights based on the \textsc{AMO}-Bench experimental findings.

\paragraph{The model output length could indicate the reasoning challenge of the benchmark.}
Section~\ref{sec:data_stat} provides a pre-analysis of benchmark difficulty based on annotated solution lengths.
Here, we offer a post-hoc analysis of benchmark difficulty based on the relationship between model performance and model output length.
Figure~\ref{fig:bmk_tokens} clearly demonstrates that the average output length of each model increases as the reasoning benchmark becomes more challenging.
Specifically, across six models, benchmarks with higher accuracy scores (such as MAH500 and AIME24) correspond to shorter average outputs, while those with lower scores (like AMO-Bench) require significantly longer responses.
This suggests that harder benchmarks demand more elaborate reasoning steps or explanations from the models, resulting in increased token usage.
These results demonstrate that the model output length could be an indicator of reasoning challenge in the benchmark.

\paragraph{Performance on \textsc{AMO}-Bench still benefits from test-time scaling.}
The reasoning efficiency results discussed above indicate a correlation between model performance and output length.
Here, we conduct a more rigorous analysis by directly controlling the reasoning effort for the same model.
As shown in the Figure~\ref{fig:reasoning_effort}, all three models (GPT-5, o4-mini, and o3-mini) exhibit a near-linear growth trend in AVG@32 as the logarithm of average output length increases.
Such a trend is highly aligned with earlier experimental observations from existing benchmarks such as MATH500 and AIME24~\citep{muennighoffs1}.
This indicates that further increasing the inference budget will further drive improvements on \textsc{AMO}-Bench.

\begin{figure}[t]
    \centering
    \includegraphics[width=0.99\textwidth]{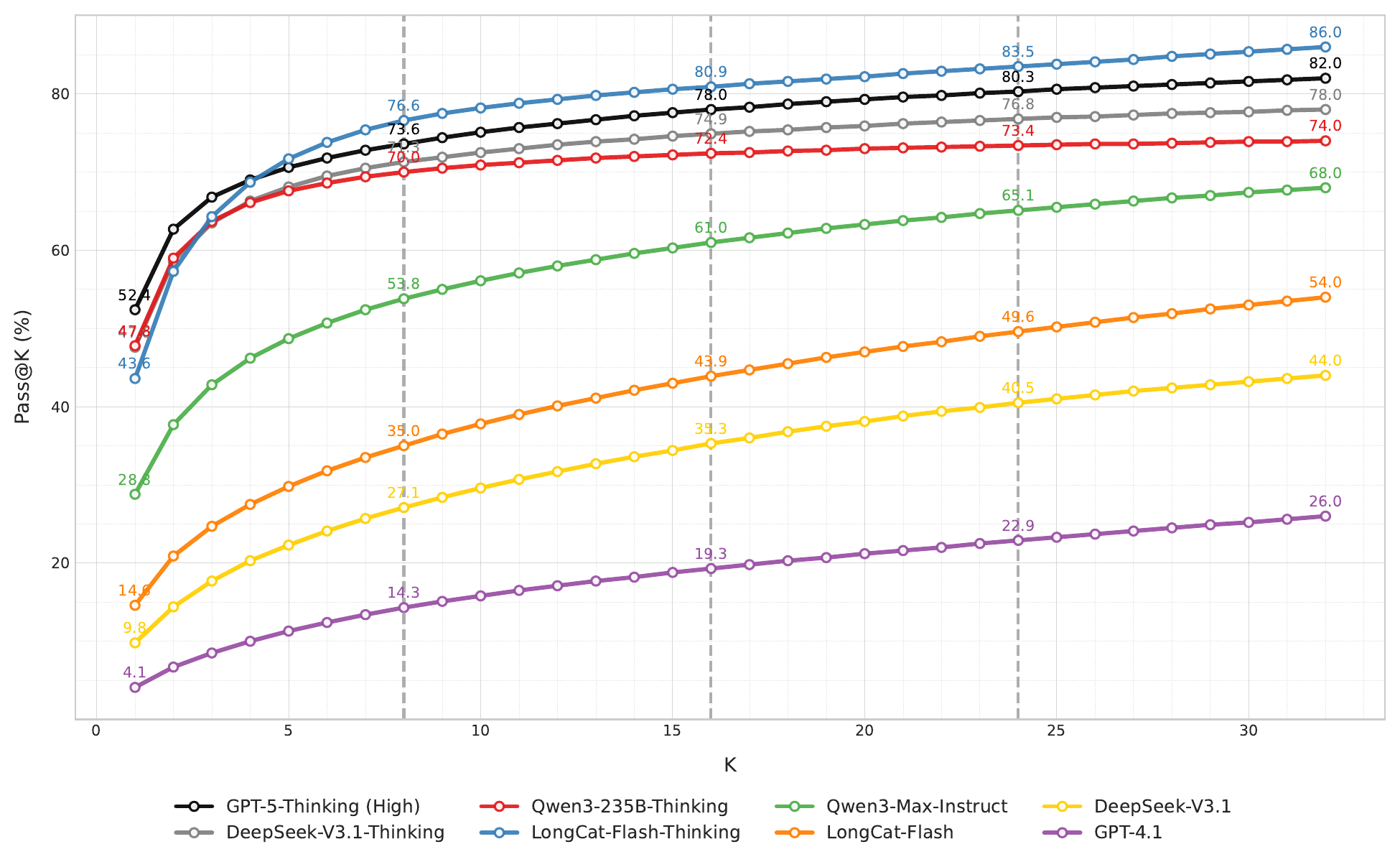}
    \caption{The the pass@$k$ trend of various LLMs with increasing $k$.}
    \label{fig:passk}
\end{figure}

\paragraph{Top-tier models demonstrate promising potential for improvement on \textsc{AMO}-Bench.}
Existing work reveals that the pass@$k$ performance of the model can reflect its inherent potential to achieve further improvement through reinforcement learning.
Inspired by this, we illustrate the pass@$k$ of evaluated models to indicate their inner potential.
As shown in Figure~\ref{fig:passk}, the pass@$k$ metric exhibits rapid growth as $k$ increases from 1 to 8, followed by a sustained but gradual improvement as $k$ continues to rise.
Notably, the top-tier reasoning models achieve over 70\% performance on the pass@32 metric.
These results highlight the significant room for improvement in the reasoning capabilities of LLMs.

\section{Related Work}

Evaluating LLMs on mathematical problem solving has been a critical aspect for assessing advancements in reasoning capabilities.
Early datasets such as GSM8K~\citep{cobbe2021training} and MATH~\citep{hendrycks2measuring} provided initial explorations to evaluate these abilities.
However, model performance on these benchmarks has quickly reached saturation.
To further advance the study of mathematical proficiency in LLMs, recent work has shifted toward more challenging benchmarks.

In terms of increasing difficulty, two primary lines of work have emerged.
One line focuses on Mathematical Olympiad (MO)-level problems, which rely on a specific range of math knowledge and require complex and intuitive reasoning skills.
For instance, Omni-MATH~\citep{gaoomni} introduces a multi-subject evaluation suite designed to rigorously test mathematical reasoning and generalization; 
OlympiadBench~\citep{he2024olympiadbench} focuses on evaluating the bilingual and multi-modal reasoning abilities with Olympid-level challenges; 
OlymMATH~\citep{sun2025challenging} collects MO-level problems from printed publications and evaluates mathematical reasoning by offering problems of two difficulty levels; 
MathOdyssey~\citep{fang2025mathodyssey} broadens the scope to include more complex tasks, with a particular focus on long-range and compositional reasoning;
BeyondAIME~\citep{beyondaime} collects problems similar in style to AIME with increased difficulty and expanded data scale;
MathArena~\citep{balunovic2025matharena} rapidly tracks model performance in newly held MO-level competitions and explores evaluation paradigms for proof-based competitions such as the IMO and USAMO.
Our proposed \textsc{AMO}-Bench also falls within this category and it stands as one of the most challenging benchmarks at the time of writing.

The other line of work focuses on problems derived from graduate-level examinations or advanced mathematical research.
For instance, RealMath~\citep{zhang2025realmath} provides a comprehensive evaluation of LLMs in real-world mathematical tasks, assessing their reasoning capabilities across a diverse range of research-level content; 
FrontierMath~\citep{glazer2024frontiermath} covers computationally intensive problems and abstract questions across most branches of mathematics, highlighting the significant gap between LLMs and the prowess of the mathematical community; 
HARDMath2~\citep{roggeveen2025hardmath2} focuses on approximation-based mathematical problems, particularly those commonly encountered in applied sciences and engineering; 
HLE~\citep{phan2025humanity} constructs a final closed-ended academic benchmark spanning multiple subjects, evaluating reasoning capabilities on human frontier knowledge.
Beside requiring the reasoning abilities, these datasets also challenge models by demanding extensive and deep mathematical knowledge.

\section{Conclusion}
We introduce \textsc{AMO}-Bench, an advanced mathematical reasoning benchmark featuring problems at the level of mathematical Olympiads or higher.
The benchmark consists of 50 human-crafted questions designed to rigorously assess advanced mathematical reasoning.
Compared with existing benchmarks, \textsc{AMO}-Bench offers more challenging assessments by ensuring that all 50 problems are entirely original and meet or exceed IMO difficulty standards.
Each problem in \textsc{AMO}-Bench requires only a final answer rather than a full proof, enabling automatic and robust grading for evaluation purposes.
Experimental results across various LLMs demonstrate that contemporary LLMs still struggle with the significant challenges presented by \textsc{AMO}-Bench.
Despite these low performances, our further analysis underscore substantial opportunities for advancing mathematical reasoning capabilities in current LLMs.

\section*{Acknowledgments}
We thank Zijian Zhang, Jun Kuang, Yiyang Li, Siyu Ren, Zongyu Wang, Yaoming Zhu, Ziyi Zhao, Linsen Guo, Yuhuai Wei, Cunguang Wang, Jiaming Wang and Mengjie Cao for their insightful suggestions regarding the construction and analysis of \textsc{AMO}-Bench.
We are grateful to Wei Wang, Wenjie Shi, Jiaqi Zhang, Xiangyu Xi, Xiangzhou Huang, Rongxiang Weng, and Jingang Wang for the valuable discussions and insights on model performance.
We also appreciate the engineering support provided by Yunke Zhao and Dengchang Zhao, and open-source assistance from Qi Li, Peng Wang and Xiangyang Ji.

\bibliographystyle{unsrtnat}
\bibliography{references}

\begin{thebibliography}{28}
\providecommand{\natexlab}[1]{#1}
\providecommand{\url}[1]{\texttt{#1}}
\expandafter\ifx\csname urlstyle\endcsname\relax
  \providecommand{\doi}[1]{doi: #1}\else
  \providecommand{\doi}{doi: \begingroup \urlstyle{rm}\Url}\fi

\bibitem[{Meituan LongCat Team}(2025{\natexlab{a}})]{longcatflashthinking}
{Meituan LongCat Team}.
\newblock Longcat-flash-thinking technical report, 2025{\natexlab{a}}.
\newblock URL \url{https://arxiv.org/abs/2509.18883}.

\bibitem[OpenAI(2024)]{o1}
OpenAI.
\newblock Openai o1 system card, 2024.
\newblock URL \url{https://arxiv.org/abs/2412.16720}.

\bibitem[{Gemini Team}(2025)]{gemini25}
{Gemini Team}.
\newblock Gemini 2.5: Pushing the frontier with advanced reasoning, multimodality, long context, and next generation agentic capabilities, 2025.
\newblock URL \url{https://arxiv.org/abs/2507.06261}.

\bibitem[OpenAI(2025)]{gpt5}
OpenAI.
\newblock Gpt-5 system card, 2025.
\newblock URL \url{https://cdn.openai.com/gpt-5-system-card.pdf}.

\bibitem[Anthropic(2025)]{claude4}
Anthropic.
\newblock System card: Claude opus 4 and claude sonnet 4, 2025.
\newblock URL \url{https://www-cdn.anthropic.com/4263b940cabb546aa0e3283f35b686f4f3b2ff47.pdf}.

\bibitem[xAI(2025)]{grok4}
xAI.
\newblock Grok 4 model card, 2025.
\newblock URL \url{https://data.x.ai/2025-08-20-grok-4-model-card.pdf}.

\bibitem[Yang et~al.(2025)Yang, Li, Yang, Zhang, Hui, Zheng, Yu, Gao, Huang, Lv, Zheng, Liu, Zhou, Huang, Hu, Ge, Wei, Lin, Tang, Yang, Tu, Zhang, Yang, Yang, Zhou, Zhou, Lin, Dang, Bao, Yang, Yu, Deng, Li, Xue, Li, Zhang, Wang, Zhu, Men, Gao, Liu, Luo, Li, Tang, Yin, Ren, Wang, Zhang, Ren, Fan, Su, Zhang, Zhang, Wan, Liu, Wang, Cui, Zhang, Zhou, and Qiu]{qwen3}
An~Yang, Anfeng Li, Baosong Yang, Beichen Zhang, Binyuan Hui, Bo~Zheng, Bowen Yu, Chang Gao, Chengen Huang, Chenxu Lv, Chujie Zheng, Dayiheng Liu, Fan Zhou, Fei Huang, Feng Hu, Hao Ge, Haoran Wei, Huan Lin, Jialong Tang, Jian Yang, Jianhong Tu, Jianwei Zhang, Jianxin Yang, Jiaxi Yang, Jing Zhou, Jingren Zhou, Junyang Lin, Kai Dang, Keqin Bao, Kexin Yang, Le~Yu, Lianghao Deng, Mei Li, Mingfeng Xue, Mingze Li, Pei Zhang, Peng Wang, Qin Zhu, Rui Men, Ruize Gao, Shixuan Liu, Shuang Luo, Tianhao Li, Tianyi Tang, Wenbiao Yin, Xingzhang Ren, Xinyu Wang, Xinyu Zhang, Xuancheng Ren, Yang Fan, Yang Su, Yichang Zhang, Yinger Zhang, Yu~Wan, Yuqiong Liu, Zekun Wang, Zeyu Cui, Zhenru Zhang, Zhipeng Zhou, and Zihan Qiu.
\newblock Qwen3 technical report, 2025.
\newblock URL \url{https://arxiv.org/abs/2505.09388}.

\bibitem[Guo et~al.(2025)Guo, Yang, Zhang, Song, Wang, Zhu, Xu, Zhang, Ma, Bi, et~al.]{deepseekr1}
Daya Guo, Dejian Yang, Haowei Zhang, Junxiao Song, Peiyi Wang, Qihao Zhu, Runxin Xu, Ruoyu Zhang, Shirong Ma, Xiao Bi, et~al.
\newblock Deepseek-r1 incentivizes reasoning in llms through reinforcement learning.
\newblock \emph{Nature}, 645\penalty0 (8081):\penalty0 633--638, 2025.

\bibitem[DeepSeek-AI(2025)]{deepseekv3}
DeepSeek-AI.
\newblock Deepseek-v3 technical report, 2025.
\newblock URL \url{https://arxiv.org/abs/2412.19437}.

\bibitem[{Meituan LongCat Team}(2025{\natexlab{b}})]{longcatflash}
{Meituan LongCat Team}.
\newblock Longcat-flash technical report, 2025{\natexlab{b}}.
\newblock URL \url{https://arxiv.org/abs/2509.01322}.

\bibitem[{GLM-4.5 Team}(2025)]{glm}
{GLM-4.5 Team}.
\newblock Glm-4.5: Agentic, reasoning, and coding (arc) foundation models, 2025.
\newblock URL \url{https://arxiv.org/abs/2508.06471}.

\bibitem[{ByteDance Seed}(2025)]{seed15}
{ByteDance Seed}.
\newblock Seed1.5-thinking: Advancing superb reasoning models with reinforcement learning, 2025.
\newblock URL \url{https://arxiv.org/abs/2504.13914}.

\bibitem[{Tencent Hunyuan Team}(2025)]{hunyuan}
{Tencent Hunyuan Team}.
\newblock Hunyuan-turbos: Advancing large language models through mamba-transformer synergy and adaptive chain-of-thought, 2025.
\newblock URL \url{https://arxiv.org/abs/2505.15431}.

\bibitem[{Kimi Team}(2025)]{kimik2}
{Kimi Team}.
\newblock Kimi k2: Open agentic intelligence, 2025.
\newblock URL \url{https://arxiv.org/abs/2507.20534}.

\bibitem[Balunovi{\'c} et~al.(2025)Balunovi{\'c}, Dekoninck, Petrov, Jovanovi{\'c}, and Vechev]{balunovic2025matharena}
Mislav Balunovi{\'c}, Jasper Dekoninck, Ivo Petrov, Nikola Jovanovi{\'c}, and Martin Vechev.
\newblock Matharena: Evaluating llms on uncontaminated math competitions.
\newblock \emph{arXiv preprint arXiv:2505.23281}, 2025.

\bibitem[He et~al.(2024)He, Luo, Bai, Hu, Thai, Shen, Hu, Han, Huang, Zhang, et~al.]{he2024olympiadbench}
Chaoqun He, Renjie Luo, Yuzhuo Bai, Shengding Hu, Zhen Thai, Junhao Shen, Jinyi Hu, Xu~Han, Yujie Huang, Yuxiang Zhang, et~al.
\newblock Olympiadbench: A challenging benchmark for promoting agi with olympiad-level bilingual multimodal scientific problems.
\newblock In \emph{Proceedings of the 62nd Annual Meeting of the Association for Computational Linguistics (Volume 1: Long Papers)}, pages 3828--3850, 2024.

\bibitem[Gao et~al.(2024)Gao, Song, Yang, Cai, Miao, Dong, Li, Ma, Chen, Xu, et~al.]{gaoomni}
Bofei Gao, Feifan Song, Zhe Yang, Zefan Cai, Yibo Miao, Qingxiu Dong, Lei Li, Chenghao Ma, Liang Chen, Runxin Xu, et~al.
\newblock Omni-math: A universal olympiad level mathematic benchmark for large language models.
\newblock In \emph{The Thirteenth International Conference on Learning Representations}, 2024.

\bibitem[Fang et~al.(2025)Fang, Wan, Lu, Xing, and Zou]{fang2025mathodyssey}
Meng Fang, Xiangpeng Wan, Fei Lu, Fei Xing, and Kai Zou.
\newblock Mathodyssey: Benchmarking mathematical problem-solving skills in large language models using odyssey math data.
\newblock \emph{Scientific Data}, 12\penalty0 (1):\penalty0 1392, 2025.

\bibitem[Sun et~al.(2025)Sun, Min, Chen, Zhao, Fang, Liu, Wang, and Wen]{sun2025challenging}
Haoxiang Sun, Yingqian Min, Zhipeng Chen, Wayne~Xin Zhao, Lei Fang, Zheng Liu, Zhongyuan Wang, and Ji-Rong Wen.
\newblock Challenging the boundaries of reasoning: An olympiad-level math benchmark for large language models.
\newblock \emph{arXiv preprint arXiv:2503.21380}, 2025.

\bibitem[Petrov et~al.(2025)Petrov, Dekoninck, Baltadzhiev, Drencheva, Minchev, Balunović, Jovanović, and Vechev]{usamo}
Ivo Petrov, Jasper Dekoninck, Lyuben Baltadzhiev, Maria Drencheva, Kristian Minchev, Mislav Balunović, Nikola Jovanović, and Martin Vechev.
\newblock Proof or bluff? evaluating llms on 2025 usa math olympiad, 2025.
\newblock URL \url{https://arxiv.org/abs/2503.21934}.

\bibitem[Muennighoff et~al.(2025)Muennighoff, Yang, Shi, Li, Fei-Fei, Hajishirzi, Zettlemoyer, Liang, Candes, and Hashimoto]{muennighoffs1}
Niklas Muennighoff, Zitong Yang, Weijia Shi, Xiang~Lisa Li, Li~Fei-Fei, Hannaneh Hajishirzi, Luke Zettlemoyer, Percy Liang, Emmanuel Candes, and Tatsunori Hashimoto.
\newblock s1: Simple test-time scaling.
\newblock In \emph{Workshop on Reasoning and Planning for Large Language Models}, 2025.

\bibitem[Cobbe et~al.(2021)Cobbe, Kosaraju, Bavarian, Chen, Jun, Kaiser, Plappert, Tworek, Hilton, Nakano, et~al.]{cobbe2021training}
Karl Cobbe, Vineet Kosaraju, Mohammad Bavarian, Mark Chen, Heewoo Jun, Lukasz Kaiser, Matthias Plappert, Jerry Tworek, Jacob Hilton, Reiichiro Nakano, et~al.
\newblock Training verifiers to solve math word problems.
\newblock \emph{arXiv preprint arXiv:2110.14168}, 2021.

\bibitem[Hendrycks et~al.(2021)Hendrycks, Burns, Kadavath, Arora, Basart, Tang, Song, and Steinhardt]{hendrycks2measuring}
Dan Hendrycks, Collin Burns, Saurav Kadavath, Akul Arora, Steven Basart, Eric Tang, Dawn Song, and Jacob Steinhardt.
\newblock Measuring mathematical problem solving with the math dataset.
\newblock In \emph{Thirty-fifth Conference on Neural Information Processing Systems Datasets and Benchmarks Track (Round 2)}, 2021.

\bibitem[ByteDance-Seed(2025)]{beyondaime}
ByteDance-Seed.
\newblock Beyondaime: Advancing math reasoning evaluation beyond high school olympiads, 2025.
\newblock URL \url{https://huggingface.co/datasets/ByteDance-Seed/BeyondAIME}.

\bibitem[Zhang et~al.(2025)Zhang, Petrui, Nikoli{\'c}, and Tram{\`e}r]{zhang2025realmath}
Jie Zhang, Cezara Petrui, Kristina Nikoli{\'c}, and Florian Tram{\`e}r.
\newblock Realmath: A continuous benchmark for evaluating language models on research-level mathematics.
\newblock \emph{arXiv preprint arXiv:2505.12575}, 2025.

\bibitem[Glazer et~al.(2024)Glazer, Erdil, Besiroglu, Chicharro, Chen, Gunning, Olsson, Denain, Ho, Santos, et~al.]{glazer2024frontiermath}
Elliot Glazer, Ege Erdil, Tamay Besiroglu, Diego Chicharro, Evan Chen, Alex Gunning, Caroline~Falkman Olsson, Jean-Stanislas Denain, Anson Ho, Emily de~Oliveira Santos, et~al.
\newblock Frontiermath: A benchmark for evaluating advanced mathematical reasoning in ai.
\newblock \emph{arXiv preprint arXiv:2411.04872}, 2024.

\bibitem[Roggeveen et~al.(2025)Roggeveen, Wang, Flintoft, Donets, Nathwani, Gutierrez, Ettel, Graf, Dandavate, Nageswaran, et~al.]{roggeveen2025hardmath2}
James~V Roggeveen, Erik~Y Wang, Will Flintoft, Peter Donets, Lucy~S Nathwani, Nickholas Gutierrez, David Ettel, Anton~Marius Graf, Siddharth Dandavate, Arjun Nageswaran, et~al.
\newblock Hardmath2: A benchmark for applied mathematics built by students as part of a graduate class.
\newblock \emph{arXiv preprint arXiv:2505.11774}, 2025.

\bibitem[Phan et~al.(2025)Phan, Gatti, Han, Li, Hu, Zhang, Zhang, Shaaban, Ling, Shi, et~al.]{phan2025humanity}
Long Phan, Alice Gatti, Ziwen Han, Nathaniel Li, Josephina Hu, Hugh Zhang, Chen Bo~Calvin Zhang, Mohamed Shaaban, John Ling, Sean Shi, et~al.
\newblock Humanity's last exam.
\newblock \emph{arXiv preprint arXiv:2501.14249}, 2025.

\end{thebibliography}

\clearpage

\appendix
\section{Prompt Templates}\label{sec:prompts}
\paragraph{Query prompt template.}
In order to guide LLMs in generating answers in a parser-readable format, we use the following prompt template guide the model generation. 
There are mainly three requirements in the instruction:
the answer prefix (i.e., \texttt{\#\#\# The final answer is:}), the LaTeX box environment (i.e., \texttt{\textbackslash boxed\{\}}), and the precision requirement.
\begin{exmp}{Query Prompt Template}{query_prompt}
    ...
    
    After solving the above problem, please output your final answer in the following format:
    
    \#\#\# The final answer is:
    \$\textbackslash boxed\{\textless your answer\textgreater\}\$
    
    Example:
    
    \#\#\# The final answer is:
    \$\textbackslash boxed\{123\}\$

    The final answer should be given as precisely as possible (using LaTeX symbols such as \textbackslash sqrt, \textbackslash frac, \textbackslash pi, etc.).
    If the final answer involves a decimal approximation, it must be accurate to at least four decimal places.
\end{exmp}

\paragraph{Grading prompt template.}
We employ the LLM-based grading using o4-mini (Low) as the grading model, and use the following grading prompt to verify the equivalence between the LLM output and the reference answer.

\begin{exmp}{Grading Prompt Template}{grading_prompt}
For the following math problem, we have the reference answer and the student's answer.

Determine whether the student's answer is equivalent to the reference answer.

If equivalent, output "Correct".

If not equivalent, output "Incorrect".\\

\#\#\# Problem

...\\

\#\#\# Reference Answer

...\\

\#\#\# Student Answer

...\\

Now, please provide your judgment.

Please strictly follow the format below to summarize your conclusion at the end of your judgment:

\#\#\# Conclusion: Correct/Incorrect

If the answer involves a decimal approximation, it must be accurate to at least four decimal places.
\end{exmp}

\section{Analysis of AVG@\textit{k}}\label{sec:sample_num}
\begin{figure}[t]
    \centering
    \includegraphics[width=0.99\textwidth]{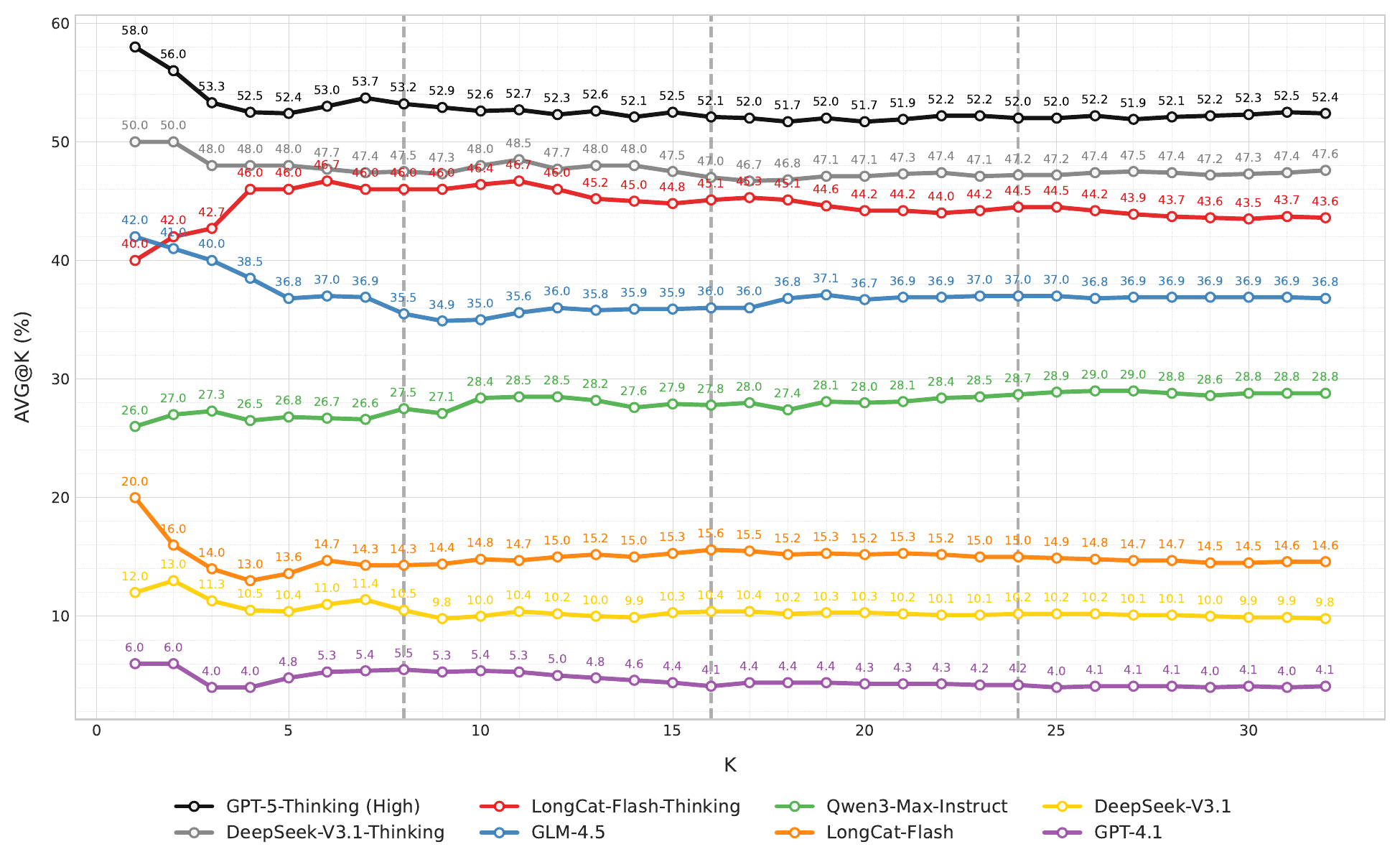}
    \caption{The AVG@$k$ trend of various LLMs with increasing $k$.}
    \label{fig:AVG@K}
\end{figure}

Figure~\ref{fig:AVG@K} illustrates the fluctuation of the average performance across different sampling times. 
It shows that as the sampling time grows, the models' performance become more stable.
When sampling 32 times, it rarely appears the reverse-order phenomenon.

\section{Performance on \textsc{AMO}-Bench-P Subset}\label{sec:amo_bench_p}
To facilitate easier reproduction and use of \textsc{AMO}-Bench, you can utilize the \textsc{AMO}-Bench-P subset, which includes only the 39 parser-based grading problems from \textsc{AMO}-Bench.
Table~\ref{tab:amo_bench_p} presents the AVG@32 performance of LLMs on \textsc{AMO}-Bench-P.
In general, performance on \textsc{AMO}-Bench-P tends to be slightly higher than on the full \textsc{AMO}-Bench, as problems requiring complex descriptive answers are inherently more challenging than those with simple-format answers.

\begin{table*}[t]
\renewcommand\arraystretch{1.2}
\Huge
\caption{The AVG@32 performance of LLMs on the \textsc{AMO}-Bench and \textsc{AMO}-Bench-P, the latter of which contains only 39 parser-based grading problems.
}
\label{tab:amo_bench_p}
\centering
\resizebox{.6\linewidth}{!}{
\begin{tabular}{@{}ccc@{}}
\toprule
Model                         & AMO-Bench & AMO-Bench-P \\ \midrule
GPT-5-Thinking (High)         & 52.4      & 54.8        \\
Qwen3-235B-A22B-Thinking-2507 & 47.8      & 56.2        \\
DeepSeek-V3.1-Thinking        & 47.6      & 53.0        \\
LongCat-Flash-Thinking        & 43.6      & 45.3        \\
o4-mini (High)                & 40.2      & 43.8        \\
Gemini-2.5-Pro                & 38.7      & 41.7        \\
GLM-4.5                       & 36.8      & 41.0        \\
Qwen3-Next-80B-Thinking       & 34.8      & 37.4        \\
DeepSeek-R1-0528              & 34.3      & 37.1        \\
o3-mini (High)                & 32.3      & 34.0        \\
Qwen3-Max-Instruct            & 28.8      & 30.9        \\
Qwen3-Next-80B-Instruct       & 18.2      & 17.8        \\
Gemini-2.5-Flash              & 18.1      & 18.0        \\
Claude-Sonnet-4.5             & 17.6      & 18.1        \\
LongCat-Flash                 & 14.6      & 14.9        \\
DeepSeek-R1                   & 10.9      & 11.7        \\
Claude-Opus-4                 & 10.6      & 11.4        \\
DeepSeek-V3.1                 & 9.8       & 9.6         \\
Kimi-K2                       & 7.5       & 8.4         \\
DeepSeek-V3-0324              & 5.2       & 5.4         \\
GPT-4.1                       & 4.1       & 4.8         \\
GPT-4o-20241120               & 1.5       & 1.9         \\ \bottomrule
\end{tabular}
}
\end{table*}

\end{document}